\documentclass[conference]{IEEEtran}
\IEEEoverridecommandlockouts
\usepackage{float}
\usepackage{cite}
\usepackage{graphicx}
\usepackage{amsmath,amssymb,amsfonts}
\usepackage{url}
\usepackage{algorithmic}
\usepackage{textcomp}
\usepackage{xcolor}
\usepackage{tabularx} % For automatic text wrapping
\usepackage{array}    % For custom column definitions
\usepackage{booktabs} % For professional-looking tables
\usepackage{graphicx}
\usepackage{makecell} % Required for multi-line headers
\usepackage{amssymb}
\usepackage{listings}
\usepackage{adjustbox}
\usepackage[autostyle]{csquotes}
\usepackage[activate=false,kerning]{microtype}
\usepackage[colorlinks=true, linkcolor=blue, urlcolor=blue, citecolor=black]{hyperref}
\usepackage[square,numbers]{natbib}
\usepackage{multirow}
\usepackage{threeparttable}
\usepackage{pifont}
\usepackage{tabularx}
\usepackage{pifont} % Required for \ding
\usepackage{colortbl} % Required for \rowcolor

\def\BibTeX{{\rm B\kern-.05em{\sc i\kern-.025em b}\kern-.08em
    T\kern-.1667em\lower.7ex\hbox{E}\kern-.125emX}}
\begin{document}

\newcommand{\perfup}[1]{\textcolor{green!70!black}{$\uparrow$ #1}}
\newcommand{\perfdown}[1]{\textcolor{red}{$\downarrow$ #1}}

\definecolor{codegray}{rgb}{0.5,0.5,0.5}
\definecolor{backcolour}{rgb}{0.95,0.95,0.92}

\lstset{
    backgroundcolor=\color{backcolour},
    basicstyle=\ttfamily\footnotesize,
    breaklines=true,
    captionpos=b,
    keepspaces=true,
    numbers=left,
    numberstyle=\tiny\color{codegray},
    showspaces=false,
    frame=single
}
% first define some new commands
%
\newcommand{\squishlist}{
\begin{list}{$\bullet$}
{   \setlength{\itemsep}{0pt}
   \setlength{\parsep}{3pt}
   \setlength{\topsep}{3pt}
   \setlength{\partopsep}{0pt}
   \setlength{\leftmargin}{1.5em}
   \setlength{\labelwidth}{1em}
   \setlength{\labelsep}{0.5em} } }
\newcounter{Lcount}
\newcommand{\squishlisttwo}{
\begin{list}{\arabic{Lcount}. }
  { \usecounter{Lcount}
 \setlength{\itemsep}{0pt}
 \setlength{\parsep}{0pt}
 \setlength{\topsep}{0pt}
 \setlength{\partopsep}{0pt}
 \setlength{\leftmargin}{2em}
 \setlength{\labelwidth}{1.5em}
 \setlength{\labelsep}{0.5em} } }
\newcommand{\squishend}{\end{list} }

\title{CircuitLM: A Multi-Agent LLM-Aided Design Framework for Generating Circuit Schematics from Natural Language Prompts}

\author{
    \IEEEauthorblockN{Khandakar Shakib Al Hasan$^1$, Syed Rifat Raiyan$^1$, Hasin Mahtab Alvee$^1$, Wahid Sadik$^2$}
    \IEEEauthorblockA{$^1$\textit{Department of Computer Science and Engineering}, $^2$\textit{Department of Electrical and Electronic Engineering} \\
    \textit{Islamic University of Technology}\\
    Dhaka, Bangladesh \\
    \{shakibalhasan, rifatraiyan, hasinmahtab, wahidsadik\}@iut-dhaka.edu}
}
% \author{
% \IEEEauthorblockN{Anonymous Author(s)}
% \IEEEauthorblockA{Affiliation Omitted for Double-Blind Review}
% }

% \and
% \and
% \and
% \and
% \and

\maketitle
% \IEEEpeerreviewmaketitle

\pagenumbering{arabic}

\begin{abstract}
Generating accurate circuit schematics from high-level natural language descriptions remains a persistent challenge in electronic design automation (EDA), as large language models (LLMs) frequently hallucinate components, violate strict physical constraints, and produce non-machine-readable outputs. To address this, we present \textbf{CircuitLM}, a multi-agent pipeline that translates user prompts into structured, visually interpretable \texttt{CircuitJSON} schematics. The framework mitigates hallucination and ensures physical viability by grounding generation in a curated, embedding-powered component knowledge base through five sequential stages: (i) component identification, (ii) canonical pinout retrieval, (iii) chain-of-thought reasoning, (iv) JSON schematic synthesis, and (v) interactive force-directed visualization. We evaluate the system on a dataset of 100 unique circuit-design prompts using five state-of-the-art LLMs. To systematically assess performance, we deploy a rigorous dual-layered evaluation methodology: a deterministic Electrical Rule Checking (ERC) engine categorizes topological faults by strict severity (Critical, Major, Minor, Warning), while an LLM-as-a-judge meta-evaluator identifies complex, context-aware design flaws that bypass standard rule-based checkers. Ultimately, this work demonstrates how targeted retrieval combined with deterministic and semantic verification can bridge natural language to structurally viable, schematic-ready hardware and safe circuit prototyping. Our code and data will be made public.
\end{abstract}


\section{Introduction}
\begin{figure}[t]
\centering
        \includegraphics[width=0.8\columnwidth]{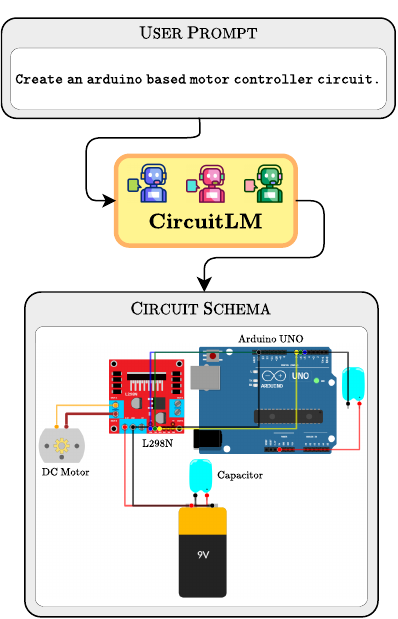}
        % \vspace{-7mm}
        \caption{Example of a circuit schema generated by CircuitLM from one of our benchmark prompts.}
        % \vspace{-3mm}
        \label{fig:overview}
\end{figure}

Electronics design requires deep expertise, precise component knowledge, and manual schematic drafting—barriers that exclude novices and rapid prototypers. Recent LLMs promise natural-language circuit design (\textit{e.g.}, ``Build a gesture-control circuit with an ESP32-C3 Mini and five MPU6050 sensors''), yet in practice, they generate unstructured textual wiring descriptions that are ambiguous, incomplete, and electrically unsound. Connections are often inaccurate, essential supporting components (pull-ups, decoupling capacitors, current-limiting resistors, $\text{I$^2$C}$ multiplexers) are omitted, and power distribution constraints are ignored. 

Crucially, these prose outputs are incompatible with standard EDA tools such as \textit{Fritzing}\footnote{\url{https://fritzing.org/}}  and \textit{Proteus Design Suite}\footnote{\url{https://www.labcenter.com/}}. Users must manually reconstruct the design in an EDA environment—a time-consuming process that frequently exposes additional logical and electrical faults. Consequently, LLM-generated circuits are rarely functional or safe on first pass.

Existing benchmarks (PINS100, MICRO25) \cite{jansen2023wordswiresgeneratingfunctioning} lack (1) holistic electrical validity checks, (2) structured machine-readable outputs, and (3) automated visualization.

To address this limitation, we introduce \textbf{CircuitLM}, a multi-agent framework that converts natural-language prompts into reliable, executable circuit schematics. Grounded in a local vector database of canonical production components, CircuitLM enforces strict pin canonicalization and invariant wiring constraints  (\textit{e.g.}, SDA$\leftrightarrow$SDA consistency, mandatory current limiting) through a five-stage pipeline.
Beyond standalone design, CircuitLM can act as a circuit design assistant if paired with platforms such as Tinkercad or Wokwi. The system can accelerate prototyping and lower the barrier for users to create and iterate on circuits, reducing the time from idea to a testable design.

We evaluate CircuitLM on a curated dataset of 100 prompts against five frontier LLMs using a rigorous ERC framework, backed by expert human validation. Our methodology evaluates circuits by:
\begin{itemize}
    \item \textbf{Categorized Severity}: Deterministically classifying topological and logical faults into Critical, Major, Minor, and Warning tiers, mirroring industry-standard EDA tools.
    \item \textbf{Pass@1 Success Rates}: Establishing absolute ground truth via human experts to define a strict success criterion (0 Critical or Major errors) for prototyping-ready schematics.
\end{itemize}

Our contributions are fourfold. First, we present CircuitLM, a multi-agent pipeline for automated circuit generation and interactive visualization from natural language. Second, we introduce \texttt{CircuitJSON}, an open, tool-agnostic schematic interchange format for structured hardware representation. Third, we establish a deterministic ERC evaluation framework with a 100-prompt benchmark dataset to assess the physical validity of LLM-generated circuits. Finally, we develop a retrieval-augmented architecture that eliminates component hallucination. By decoupling the prompt context from the total database size ($N$), the framework scales to industrial libraries with $\mathcal{O}(k)$ prompt complexity, where $k$ is the number of components required for a given design.

% \vspace{-3mm}

\section{Related Works}
Electronic circuit design automation has evolved from deterministic, heuristic-driven algorithms toward stochastic, agentic AI frameworks, reflecting the increasing complexity of embedded systems and the demand for higher-level abstraction in hardware synthesis. Early applications of Machine Learning (ML) in Electronic Design Automation (EDA) focused on localized optimization within the physical design flow, employing Convolutional Neural Networks (CNNs) for routability prediction and Graph Neural Networks (GNNs) for netlist representation and parasitic estimation \cite{talebzadeh2021machine}. While effective at specific VLSI stages, these approaches lacked the semantic capacity to interpret high-level user intent or natural language specifications. The success of LLMs in software engineering subsequently motivated research into hardware generation, particularly in Hardware Description Languages (HDLs). Systems such as VerilogEval \cite{thakur2023verilogeval} and ChipChat \cite{blocklove2023chipchat} demonstrated LLM-based RTL generation via conversational interfaces. However, generating complete electronic schematics introduces substantially higher complexity, requiring physical grounding through accurate pin mappings, power management, and integration of heterogeneous peripherals. A key challenge in autonomous schematic generation is the hallucination of connectivity, where models invent invalid pin labels or violate electrical constraints. Recent work has explored structured intermediate representations to mitigate this issue. Schemato \cite{matsuo2025schematollmnetlisttoschematic} reconstructs schematics from formal netlists, while EESchematic \cite{liu2025eeschematicmultimodalllmbasedai} investigates end-to-end generation using standardized symbol libraries. Nonetheless, existing datasets and frameworks—such as LLM4Netlist \cite{ye2025llm4netlist}—remain focused on synthesizable digital logic, leaving embedded systems and analog sensor-rich designs largely unaddressed.
To mimic human engineering workflows, research has pivoted toward Multi-Agent Systems (MAS), which have already demonstrated good efficacy in mathematical \cite{wan2025rema} and physics \cite{siddique2025physicseval} reasoning. By decomposing design into specialized roles—such as design, verification, and routing agents—MAS frameworks employ \enquote{debate and critique} mechanisms to reduce error rates \cite{hong2024metagpt}. Cutting-edge frameworks like MenTeR \cite{chen2025menterfullyautomatedmultiagentworkflow} utilize diagram-aware RAG (Retrieval-Augmented Generation) and \enquote{circuit think tanks} for automated RF/analog design. Furthermore, studies on agent topologies suggest that distributed reasoning consistently outperforms monolithic models by isolating sub-task complexity \cite{pan2025surveyresearchlargelanguage}. 

As evident in Table \ref{tab:related_works_comparison}, CircuitLM bridges these gaps by introducing a novel multi-agentic pipeline that enforces physical grounding through a canonical retrieval database. By moving beyond simple text generation to a structured, reasoning-first approach, this study provides the first comprehensive evidence that agentic frameworks can handle the multidimensional constraints of real-world embedded prototyping, producing rapid-prototyping outputs that are both logically sound and visually interpretable.

% \vspace{-2mm}

\section{Methodology}
\begin{figure}[t]
    \centering
    \includegraphics[width=0.9\columnwidth]{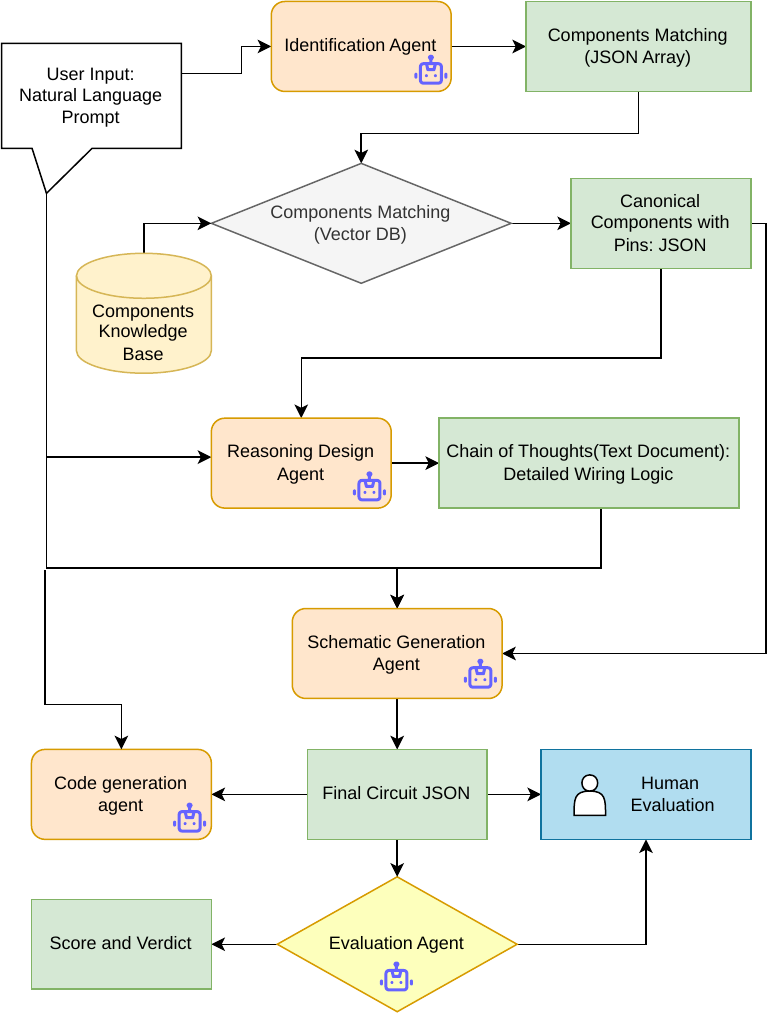}
    \caption{An overview of the CircuitLM framework.}
    \label{fig:architecture}
    % \vspace{-2mm}
\end{figure}

We propose a multi-step, multi-agent pipeline that transforms a high-level, natural language project idea (the user prompt) into a structured, logically sound circuit definition. The pipeline contains the following 5 stages:

% \vspace{-3mm}

\subsection{Stage I: Components Extraction (Identification Agent)}
In the initial phase, a specialized \textbf{Identification Agent} parses the user prompt and performs named entity recognition (NER) plus intent analysis to extract a comprehensive hardware list. To constrain the search space and reduce tokens, the agent is limited to generic component categories (\textit{e.g.}, Pressure Sensor, Arduino Uno, IMU Sensor) rather than full component records. The output is a structured JSON array of component names (\textit{e.g.}, \texttt{["Microcontroller", "Motor Driver", "DC motor"]}). \autoref{fig:compident} shows the identification agent returning the required component list.

% \vspace{-3mm}

\subsection{Stage II: Component Matching (Retrieval Agent)}
The list of Stage I is passed to the \textbf{Retrieval Agent}. The agent runs an embedding-based similarity search over a local vector knowledge base of canonical components \cite{lewis2021retrievalaugmentedgeneration}, then augments it with a fuzzy lookup using a dictionary of equivalent parts. Results are merged and mapped to records in the \texttt{component} database, yielding a dictionary of verified canonical names and precise pin-out definitions (\autoref{fig:compretrieve}). This step prevents LLM pin hallucinations by sourcing exact pin names and numbers from the curated database. If a requested part is missing, an Out-of-Distribution (OOD) flag is raised, execution halts, and human review is requested. OOD halt is a strict safety guardrail to prevent hallucinated wiring.

% \vspace{-3mm}

\subsection{Stage III: Design Reasoning (Electronics Expert Agent)}
The \textbf{Electronics Expert Agent} is the cognitive core of the pipeline. Given the user prompt, validated component list, and pin definitions, it performs high-level engineering reasoning and produces a structured Chain-of-Thought (CoT) document \cite{wei2023chainofthoughtpromptingelicitsreasoning}. The CoT explicitly encodes functional goals, power requirements, safety constraints, and pin-level wiring logic. Using least-to-most prompting \cite{zhou2023leasttomostpromptingenablescomplex}, the agent decomposes complex requirements into hierarchical sub-tasks, improving electrical logicality prior to code generation \cite{wang2023selfconsistencyimproveschainthought}. \autoref{fig:cot} shows the CoT agent inputs and outputs.

% \vspace{-3mm}

\subsection{Stage IV: Schema Generation (Circuit Generation Agent)}
The \textbf{Circuit Generation Agent} transforms the CoT document and canonicalized pin mappings into a machine-readable schematic representation (\autoref{fig:schematic}). It produces a strictly formatted \texttt{CircuitJSON} object, inspired by the Wokwi\footnote{\url{https://wokwi.com/}} simulation platform~\cite{inproceedings}. The output specifies exact component placement (coordinates and rotation) together with a complete, canonicalized list of pin-to-pin electrical connections. Each component entry additionally contains a dedicated \texttt{attrs} field for parametric configuration. This field stores structured key–value pairs representing electrical parameters (\textit{e.g.}, resistance, capacitance, inductance), operating constraints (\textit{e.g.}, voltage limits), or optional metadata required for simulation and validation. By separating structural connectivity from parametric specification, the schema ensures deterministic wiring while preserving flexibility for component-level configuration. The resulting JSON is directly consumable by the schematic viewer; \autoref{lst:schema} details the full structure.

% \vspace{-3mm}

\subsection{Stage V: Schematic Visualizer}
The browser-based visualization engine renders \texttt{CircuitJSON} into interactive SVG schematics, mapping recognized components to predefined symbols and unknown items to labeled rectangular glyphs. Initial component placement utilizes a force-directed algorithm \cite{Schönfeld2019} to group connected parts and prevent overlap, while wire routing applies a Manhattan-style strategy \cite{Lee1961AnAF} prioritizing short paths and minimal bends. Crucially, this serves as a dynamic, human-in-the-loop interface allowing users to drag components and reshape wires in real-time. (\autoref{fig:schematic} and \autoref{fig:schematic2} display the raw, unadjusted autonomous outputs, while the end-to-end pipeline is summarized in \autoref{fig:architecture}).

Diverging from traditional, netlist-heavy EDA workflows, CircuitLM's visual-first paradigm is lightweight and optimized for early prototyping and education. Because the interactive interface allows seamless manual layout resolution, strictly optimizing deterministic graph-drawing metrics (\textit{e.g.}, minimizing edge crossings) is unnecessary. Such routing optimizations remain a separate challenge outside our primary scope of evaluating LLM electrical reasoning.

% \vspace{-3mm}

\subsection{Firmware Code Generation}
While the primary pipeline focuses on circuit generation and evaluation, we include a firmware code-generation agent to demonstrate automated deployment (not part of the core workflow). For ESP32 and Arduino targets, we pass the prompt and generated \texttt{CircuitJSON} to the code agent. LLMs already perform strongly on software benchmarks (HumanEval \cite{chen2021evaluatinglargelanguagemodels}, MBPP \cite{austin2021programsynthesislargelanguage}); since embedded firmware often follows repeatable patterns, producing syntactically correct firmware is comparatively easier than hardware synthesis.

% \vspace{-3mm}

\subsection{ERC Evaluation Framework}
While circuit simulation and ERC are standard in traditional EDA workflows, legacy tools are often poorly matched for LLM-driven, board-level generation. SPICE, the gold standard for continuous-time analog analysis, is computationally infeasible and practically irrelevant for our microcontroller-centric targets (\textit{e.g.}, Arduino/ESP32), which are dominated by digital I/O and discrete firmware states. Similarly, traditional ERC engines rely on rigid, proprietary netlist formats unsuited for lightweight generative workflows. Rather than bypassing verification, we developed a custom, deterministic ERC engine engineered specifically to parse \texttt{CircuitJSON}. By modeling the generated schematic as a topological graph (via \texttt{networkx}), our framework performs programmatic pathfinding to verify physical safety and logical constraints. The engine evaluates circuits against a curated knowledge base, deterministically categorizing faults by severity (Critical, Major, Minor, Warning). The registry of automated checkers includes galvanic short-circuit detection, passive component verification (\textit{e.g.}, current-limiting resistors for LEDs, pull-up resistors), inductive load protection (flyback diodes), logic-level voltage matching, and floating input detection. This tailored, graph-based approach allows us to rigorously evaluate the physical validity of LLM-generated schematics without the overhead of continuous-time analog simulation.

% \vspace{-2mm}
\section{Experimental Setup}
The experiment was designed to ensure fair and robust benchmarking of the candidate LLMs. We run the experiments multiple times, and the final score is calculated as the mean across all runs. This setup contains the implementation of the knowledge base and the configuration of the LLM agents.

\vspace{-2mm}

\subsection{Local Knowledge Base}
A key element of the system is the curated component library, stored in a ChromaDB vector database. This database serves two primary functions:
\squishlist
    \item \textbf{Components List}: Providing a finite and authoritative set of components, including their canonical identifiers, mandatory pin labels, and known aliases. This enables strict evaluation while still supporting flexible component retrieval.
    \item \textbf{Semantic Component Resolution}: Translating verbally requested components generated by the LLM in Stage I into definitive canonical component keys and pin sets required for the final \texttt{CircuitJSON} representation in Stage IV.
    \item \textbf{Database}: We choose ChromaDB for its lightweight architecture and schema flexibility, allowing rich component metadata (pins, aliases, specifications) to be stored alongside embeddings, simplifying the mapping from semantic queries to canonical component definitions. For this experiment, we utilize a set of 70+ components representing a broad range of component types.
    This selection establishes a foundational taxonomy of standard embedded interfaces (\textit{e.g.}, I$^2$C, SPI, PWM). Because devices within the same family share identical wiring logic, this curated set comprehensively covers practical design patterns. Our primary objective is to demonstrate the framework's structural reasoning across distinct protocols; expanding the component library is simply a data-entry task.
    
    \item \mbox{\textbf{Embedding Model}}: \texttt{Qwen3-Embedding-0.6B} \cite{yang2025qwen3} model is used to compute dense vector representations of component descriptions. Component matching is performed using cosine similarity with an empirically determined similarity threshold. The Qwen3 embedding model provides high-quality semantic representations for short, technical phrases and structured component metadata in electronics domains. The model offers a favorable balance between retrieval accuracy and computational efficiency \cite{zhang2025qwen3embeddingadvancingtext}, making it well-suited for persistent vector search \cite{lewis2021retrievalaugmentedgeneration}.
    \item \textbf{Library Structure}: Each component entry includes a canonical name (key), mandatory pin labels, physical dimensions (width and height), aliases, a natural language description, category, typical usage, and technical specifications. Strict adherence to the defined pin labels (\textit{e.g.,} \texttt{D21}, \texttt{GND}, \texttt{VCC}) is enforced during final evaluation (Dimension~1).
\squishend

% \vspace{-3mm}
\subsection{Benchmarked LLMs}
A total of five distinct LLMs were benchmarked as candidate models across the key text-generation stages of the proposed multi-agent pipeline (Stages~I, III, and IV). The evaluated models are: \texttt{gpt-5-mini}~\cite{openai2025gpt5},
\texttt{gemini-2.5-flash}~\cite{team2023gemini},
\texttt{deepseek-v3.1}~\cite{liu2024deepseekv3},
\texttt{qwen3-235b-a22b-2507}~\cite{yang2025qwen3},
and \texttt{llama-3.3-70b-instruct}~\cite{touvron2024llama3}.

All models were accessed through a unified API interface to ensure consistency in prompt formatting and execution conditions, and the temperature parameter was fixed at $0$ for all experimental runs.\footnote{Minor run-to-run variance may occur despite temperature 0 due to API-level nondeterminism.} For each experimental run, the same base model was used across all agents, ensuring that model performance was evaluated consistently without mixing different models across pipeline stages.

\vspace{-3mm}

\subsection{Dataset (User Prompts)}
A prompt-based dataset for netlist generation was introduced in 2025~\cite{ye2025llm4netlist}. It focuses exclusively on synthesizable digital logic gate circuits in Verilog, excluding embedded systems, IoT, and Arduino-specific components such as PWM motor drivers and sensor interfaces. In contrast, our experiment utilizes a set of 100 diverse user prompts covering a broad range of embedded systems projects, from basic LED control and motor speed regulation to complex multi-actuator circuits, communication bus interfacing (I\textsuperscript{2}C, SPI, UART), and sensor integration (IMU, DHT, PIR, LDR, \textit{etc.}). The selection criteria focused on emulating real-world embedded systems, robotics, and IoT hardware requests, ensuring a rigorous test of the system's ability to handle complexity, ambiguity, and technical constraints. The CircuitLM benchmark thus primarily targets the embedded systems domain. To systematically assess model limitations, the dataset is stratified into three distinct difficulty tiers—Easy, Medium, and Adversarial—ensuring a balanced distribution of complexity across the benchmark. To support external validity and future reproducibility, the complete dataset, including all prompts and canonical schemas, will be open-sourced upon publication.

% \vspace{-3mm}

\subsection{Evaluation Methodology}

To evaluate the physical viability of LLM-generated schematics, we implement a deterministic \textbf{ERC} framework inspired by standard Electronic Design Automation (EDA) workflows. Functioning as a safety-first verification layer, the system scans circuits for electrical violations and produces structured error reports categorized by severity.

\textbf{Technical Implementation.} The pipeline utilizes a topological graph analyzer that parses schematics into a bipartite graph of components, pins, and galvanic nets. Passive components are modeled using weighted traversal paths to mathematically distinguish direct shorts from resistive bridges. The system strictly enforces logic-level matching ($V_{\text{out}} > V_{\text{in\_max}}$), detects floating inputs or net contention, and validates component-specific requirements (\textit{e.g.,} PWM mapping) against a curated hardware \textbf{Knowledge Base (KB)}. This KB contains constraints for maximum operating voltages, current limits, pin multiplexing, and inductive load requirements.

\textbf{Error Classification.} Each model is tested three times. Detected issues are classified into four tiers:
\begin{itemize}
    \item \textbf{Fatal (Safety-Critical):} Topological violations causing immediate damage, such as VCC–GND shorts identified via shortest-path traversal.
    \item \textbf{Major (Functional):} Faults guaranteeing system failure, including logic-level mismatches, missing current-limiting resistors, or absent flyback diodes.
    \item \textbf{Minor (Performance):} Suboptimal configurations, such as technically valid but poorly chosen passive values.
    \item \textbf{Warnings (Best Practices):} Deviations from engineering conventions, such as missing decoupling capacitors.
\end{itemize}

\textbf{Primary Metric: Pass@1.} Generative performance is measured using a binary \textbf{Pass@1} metric. A schematic is a ``Pass'' only if the ERC detects \textbf{zero Fatal and zero Major errors}. Minor errors and warnings are recorded for diagnostics but do not trigger failure. This criterion mirrors real-world standards where a single critical fault renders a board unusable. OOD halts are counted as automatic failures.

\textbf{Blind Expert Verification.} To validate the automated evaluation, three electrical engineering experts who underwent a calibration phase utilizing a strict, predefined error taxonomy, conducted a blind review of the circuits and ERC reports. Inter-rater reliability was quantified using \textbf{Fleiss' kappa ($\kappa$)} to provide a statistical measure of agreement among evaluators regarding circuit correctness and diagnostic accuracy. we extracted a stratified random sample comprising 25\% of the generated circuits from each of the five frontier models, balanced across difficulty tiers for blind manual review.

\textbf{Secondary Meta-Evaluation: LLM-as-a-Judge.}
ERC ensures electrical safety and connectivity but lacks the semantic context required to verify higher-level functionality in complex digital systems. ERC cannot guarantee the circuit works. It only checks electrical connection rules. Microcontrollers such as Arduino and ESP32 use highly multiplexed GPIOs and strict protocol requirements; thus, graph-based ERC can confirm connections and detect shorts but cannot verify higher-level logic—such as correct routing of SPI signals (MOSI/MISO), proper crossing of UART TX/RX lines, or avoiding address conflicts on I\textsuperscript{2}C buses.
To address this limitation, we introduce a meta-evaluator following the \emph{LLM-as-a-Judge} paradigm. A separate QA Agent evaluates protocol correctness, logical routing, and firmware plausibility using a different model (\texttt{Claude Sonnet 4.5}), excluded from the generator pool to avoid self-preference bias.
Errors are classified into four fixed categories same as ERC but with more generalized instructions (\textit{e.g.}, fatal errors are immediate circuit failure or damaged components, major errors mean rendering the circuit nonfunctional). The model also assumes \emph{board-level reality}, recognizing that common modules may include onboard regulators and supporting circuitry.

% \vspace{-2mm}

\begin{table*}[t]
% \vspace{-3mm}
\caption{Performance Comparison: Full-Context Zero-Shot Baseline vs. CircuitLM Pipeline}
\label{tab:zeroshot_comparison}
\centering
\resizebox{\textwidth}{!}{%
\begin{tabular}{l|cccc|cccc}
\toprule
\multirow{2}{*}{\textbf{  Model}} & \multicolumn{4}{c|}{\textbf{Zero-Shot Baseline}} & \multicolumn{4}{c}{\textbf{CircuitLM}} \\
\cmidrule(lr){2-5} \cmidrule(l){6-9}
& \textbf{ERC Pass@1} $\uparrow$ & \textbf{LLM Pass@1} $\uparrow$ & \textbf{LLM Fatal} $\downarrow$ & \textbf{LLM Major} $\downarrow$ 
& \textbf{ERC Pass@1} $\uparrow$ & \textbf{LLM Pass@1} $\uparrow$ & \textbf{LLM Fatal} $\downarrow$ & \textbf{LLM Major} $\downarrow$ \\
\midrule
Gemini 2.5 Flash & 84\% & 51\% & 0.0 $\pm$ 0.3 & 1.2 $\pm$ 1.5 & 88\% (\textcolor{green}{+4\%}) & 53\% (\textcolor{green}{+2\%}) & 0.0 $\pm$ 0.2 & 1.1 $\pm$ 1.3 \\
Qwen-3 235B      & 79\% & 33\% & 0.9 $\pm$ 0.4 & 1.7 $\pm$ 1.4 & 87\% (\textcolor{green}{+8\%}) & 38\% (\textcolor{green}{+5\%}) & 0.4 $\pm$ 0.7 & 1.8 $\pm$ 1.4 \\
Deepseek v3.1    & 77\% & 39\% & 0.5 $\pm$ 0.7 & 1.7 $\pm$ 1.3  & 86\% (\textcolor{green}{+9\%}) & 40\% (\textcolor{green}{+1\%}) & 0.3 $\pm$ 0.8 & 1.5 $\pm$ 1.6 \\
GPT-5 mini       & 85\% & 21\% & 0.1 $\pm$ 0.4 & 1.8 $\pm$ 1.2 & 83\% (\textcolor{red}{-2\%}) & 23\% (\textcolor{green}{+2\%}) & 0.1 $\pm$ 0.2 & 1.3 $\pm$ 1.4 \\
\bottomrule
\end{tabular}
}
% \vspace{-3mm}
\end{table*}

\section{Results}
\subsection{Physical Viability and Deterministic ERC Performance}
The results in \autoref{tab:erc_results} demonstrate remarkably high reliability across all models. Gemini and Qwen led with Pass@1 rates of 88\% and 87\%, respectively. A key finding is the near-total elimination of fatal errors ($\mu = 0.0$ for nearly all backends), indicating that the CoT planning phase effectively forces models to identify VCC/GND rails and avoid critical short circuits. GPT was the only model exhibiting non-zero average fatal errors ($\mu = 0.0 \pm 0.2$), though still at an extremely low frequency.
% \vspace{-3mm}
\subsection{Error Severity Analysis}
While fatal errors were rare, major errors—primarily involving logic level mismatches or missing current-limiting resistors in multi-component systems—remained the primary failure mode. Models occasionally overlooked voltage domain boundaries (\textit{e.g.,} 5V \textit{vs.} 3.3V) when interfacing heterogeneous components. Deepseek showed a unique tendency for minor errors and warnings ($\mu = 0.3 \pm 0.9$ for warnings), often flagged by the ERC for missing decoupling capacitors on active components. Inter-rater reliability analysis demonstrates near-perfect consensus between the human experts and the ERC engine in classifying Fatal and Major violations, achieving a Fleiss' kappa of $\kappa = 0.82$ (95\% CI $[0.74, 0.93]$). Agreement was near-perfect for safety-critical tiers (Fatal $\kappa=0.94$, Major $\kappa=0.88$), but slightly lower for subjective best-practices (Minor $\kappa=0.73$, Warning $\kappa=0.68$), ultimately confirming that CircuitLM reliably generates professionally viable, safe, and electrically correct hardware.

\vspace{-3mm}

\begin{table}[H]
\caption{ERC Evaluation Results: Pass@1 Rates and Errors per Prompt ($\mu \pm \sigma$)}
\label{tab:erc_results}
\centering
\resizebox{\columnwidth}{!}{%
\begin{tabular}{l|c|c|c|c|c}
\toprule
\textbf{Model} & \textbf{Pass@1} & \textbf{Fatal} & \textbf{Major} & \textbf{Minor} & \textbf{Warning} \\
\midrule
Gemini & \textbf{88\%} & 0.0 $\pm$ 0.0 & 0.1 $\pm$ 0.5 & 0.0 $\pm$ 0.0 & 0.0 $\pm$ 0.0 \\
Qwen & 87\% & 0.0 $\pm$ 0.0 & 0.2 $\pm$ 0.6 & 0.0 $\pm$ 0.0 & 0.0 $\pm$ 0.0 \\
Deepseek & 86\% & 0.0 $\pm$ 0.0 & 0.2 $\pm$ 0.5 & 0.1 $\pm$ 0.8 & 0.3 $\pm$ 0.9 \\
GPT & 83\% & 0.1 $\pm$ 0.2 & 0.3 $\pm$ 0.7 & 0.0 $\pm$ 0.0 & 0.0 $\pm$ 0.0 \\
Llama & 84\% & 0.1 $\pm$ 0.2 & 0.2 $\pm$ 0.6 & 0.0 $\pm$ 0.0 & 0.0 $\pm$ 0.0 \\
\bottomrule
\end{tabular}
}
% \vspace{-2mm}
\end{table}

% \vspace{-3mm}

\subsection{LLM-as-a-Judge Meta-Evaluation}
A significant ``evaluation gap'' is observed between the deterministic ERC Pass@1 (\autoref{tab:erc_results}) and the LLM-as-Judge Pass@1 (\autoref{tab:llm_results}). While Gemini achieved an 88\% ERC pass rate, its LLM-as-Judge pass rate dropped to 53\%. Similar trends are visible across all backends, with Llama falling from 84\% to 21\%. This discrepancy highlights the value of semantic validation in circuit generation. Blind expert audits of the LLM judge's outputs yielded a substantial Fleiss' kappa of $\kappa = 0.78$ (95\% CI $[0.65, 0.88]$), confirming the agent is not hallucinating faults, but reliably validating complex, protocol-driven hardware design. The LLM-as-Judge demonstrated superior context awareness and vast internal electrical knowledge, identifying sophisticated design flaws that bypassed the deterministic checker's static rules. \autoref{tab:difficulty_breakdown} shows the breakdown of Pass@1 scores across prompt difficulty levels for both ERC and LLM-based evaluations.
\begin{table}[H]
% \vspace{-3mm}
\caption{LLM-as-Judge Evaluation Results: Pass@1 Rates and Errors per Prompt ($\mu \pm \sigma$)}
\label{tab:llm_results}
\centering
\resizebox{\columnwidth}{!}{%
\begin{tabular}{l|c|c|c|c|c}
\toprule
\textbf{Model} & \textbf{Pass@1} & \textbf{Fatal} & \textbf{Major} & \textbf{Minor} & \textbf{Warning} \\
\midrule
Gemini & 53\% & 0.0 $\pm$ 0.2 & 1.1 $\pm$ 1.3 & 0.5 $\pm$ 0.7 & 2.5 $\pm$ 1.2 \\
Deepseek & 40\% & 0.1 $\pm$ 0.4 & 1.3 $\pm$ 1.4 & 0.5 $\pm$ 0.7 & 2.7 $\pm$ 1.5 \\
Qwen & 38\% & 0.3 $\pm$ 0.8 & 1.5 $\pm$ 1.6 & 0.5 $\pm$ 0.7 & 2.5 $\pm$ 1.2 \\
GPT & 23\% & 0.4 $\pm$ 0.7 & 1.8 $\pm$ 1.4 & 0.7 $\pm$ 0.7 & 3.3 $\pm$ 1.0 \\
Llama & 21\% & 0.4 $\pm$ 0.8 & 2.1 $\pm$ 1.6 & 0.6 $\pm$ 0.8 & 2.4 $\pm$ 1.0 \\
\bottomrule
\end{tabular}
}
% \vspace{-2mm}
\end{table}

The LLM judge frequently caught errors in digital interfaces that the ERC engine, which primarily checks for net continuity and logic levels, overlooked. This includes UART RX/TX confusion (\textit{e.g.,} connecting a microcontroller's TX pin directly to another device's TX pin) and SPI protocol miswiring or omitting required pull-up resistors on bus lines. Beyond standard GPIO, the LLM judge identified mismatches between a pin's physical capabilities and its assigned function. A recurring example was the use of \textit{non-PWM-capable pins} for pulse-width modulation tasks or the selection of multiplexed pins that were already reserved for an active SPI/UART bus, avoiding resource contention that rule-based checks might miss. While the ERC flags direct logic-level mismatches, the LLM judge identified ``brownout'' risks and \textit{undervoltage conditions} that are technically within logic tolerances but functionally invalid. For instance, attempting to drive a 5V-required inductive load (\textit{e.g.,} a 5V relay or buzzer) with a 3.3V GPIO signal was flagged as a major error, even if the ERC registry categorized the connection as a valid digital interface. \autoref{fig:eval_circuit} illustrates the errors identified by the LLM in a sample circuit.
% \vspace{-1mm}

\section{Ablation Study: Zeroshot Baseline}
To evaluate the necessity of the hybrid retrieval module and the multi-agent orchestration within the CircuitLM architecture, we conduct a baseline ablation study on the top 3 models.

% \vspace{-3mm}

\subsection{Ablation Setup}
A standard zero-shot model must either generate circuit descriptions without grounding—leading to systematic component hallucination and a failure to produce visually renderable schematics—or receive the entire component library in context. To establish the strongest possible baseline, we bypassed CircuitLM's retrieval agent and CoT agent entirely. Instead, the full, unfiltered component library, complete with all semantic descriptions and exact pin definitions, was injected directly into a single system prompt. Additionally, we run the ablation removing only the CoT agent. This configuration forced the generator LLM to select components and route connections in a single pass. To account for stochastic variation in model outputs, we ran each experiment 3 times for every model and report the average performance along with the variance.

% \vspace{-4mm}

\subsection{Ablation Results}
The two approaches diverge significantly in terms of computational overhead and architectural scalability. The zero-shot baseline consumes $\sim$15,000 tokens per request due to the exhaustive library injection. In contrast, the CircuitLM pipeline's targeted retrieval and structured planning use only $\sim$3,000 tokens per request. 
While the zero-shot baseline exhibited lower average latency ($\sim$3 seconds \textit{vs.} $\sim$8 seconds for CircuitLM due to multiple API calls), the full-context approach fundamentally fails to scale. As component libraries expand toward industrial proportions, injecting thousands of components data into a single prompt becomes cost-prohibitive and inevitably breaches the context window limits of most models. Notable in our testing was the exclusion of \textbf{Llama-3.3}, which consistently failed to generate outputs in the zero-shot setting due to immediate context overflows.

Our quantitative analysis in \autoref{tab:zeroshot_comparison} indicates that for high-parameter “reasoning” models, the performance gap is relatively small. This is largely because modern frontier models are inherently trained to perform internal Chain-of-Thought (CoT) reasoning. However, incorporating an explicit CircuitLM CoT planning phase adds a crucial safety layer, helping ensure that complex circuits are constructed correctly, as demonstrated in \autoref{tab:difficulty_breakdown_cot}.

Interestingly, we observed a performance degradation in \textbf{GPT-5-mini}. We hypothesize that for distilled, highly optimized models, overly prescriptive planning instructions may interfere with internal heuristic efficiencies. Nevertheless, for the majority of the backends, the CircuitLM pipeline provides the robust, context-efficient architecture necessary for reliable Electronic Design Automation (EDA).

% \vspace{-1mm}

\section{Limitations and Conclusion}
A notable limitation of our evaluation framework is the absence of simulator-in-the-loop verification (\textit{e.g.,} automated Wokwi or Proteus co-simulation). Limitations also include computational overhead from iterative multi-agent queries and variable latency depending on model and API traffic. LLM-based validation relied on a single evaluator. Our evaluation focused primarily on digital and embedded electronics, and we did not experimentally assess performance on analog circuits (\textit{e.g.}, op-amp–based designs). Additionally, converting \texttt{CircuitJSON} into other representations, such as a netlist, would require robust scripting and additional engineering effort, which falls outside the scope of this work.

CircuitLM introduces a multi-agent framework that translates natural-language intent into physically realizable electronic schematics. By incorporating a CoT phase and an evaluation framework (deterministic ERC and LLM-as-a-Judge) we significantly improved design reliability across multiple frontier models. Future work will focus on a heterogeneous multi-LLM evaluation system, utilizing agreement-based consensus and iterative feedback loops to validate and correct connectivity logic before synthesis. We will explore a direct transpilation layer from CircuitLM’s connection arrays into Python-based SKiDL definitions, allowing the automated generation of KiCad-compatible netlists. We also plan to explore ensemble architectures to improve scalability across complex industrial designs. Finally, we propose the \texttt{CircuitJSON} format as a standardized bridge between natural language and hardware description, encouraging its adoption for future dataset creation in the EDA community.

\bibliographystyle{IEEEtran}
\bibliography{references}

\newpage
\appendix
\section{Appendix}

\label{sec:appendix}
\begin{figure}[H]
    \centering
    \includegraphics[width=0.9\columnwidth]{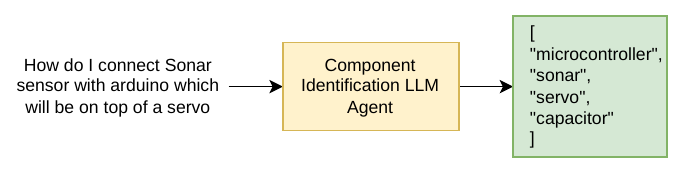}
    \caption{Input output data of component identification agent}
    \label{fig:compident}
\end{figure}

\begin{figure}[H]
    \centering
    \includegraphics[height=\columnwidth]{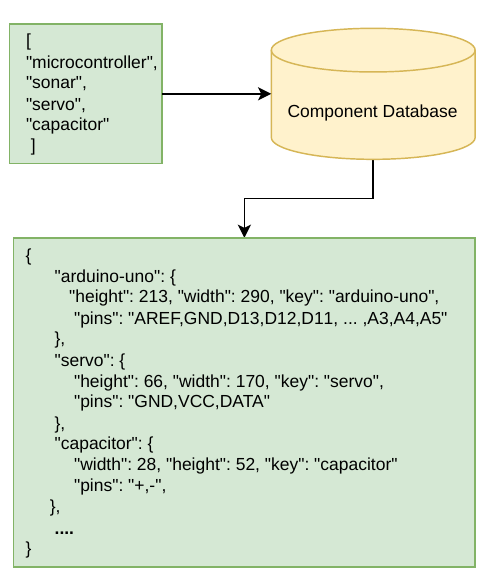}
    \caption{Components pin out information is retrieved by using the component's name from the database}
    \label{fig:compretrieve}
\end{figure}

\begin{figure}[H]
    \centering
    \includegraphics[width=\columnwidth]{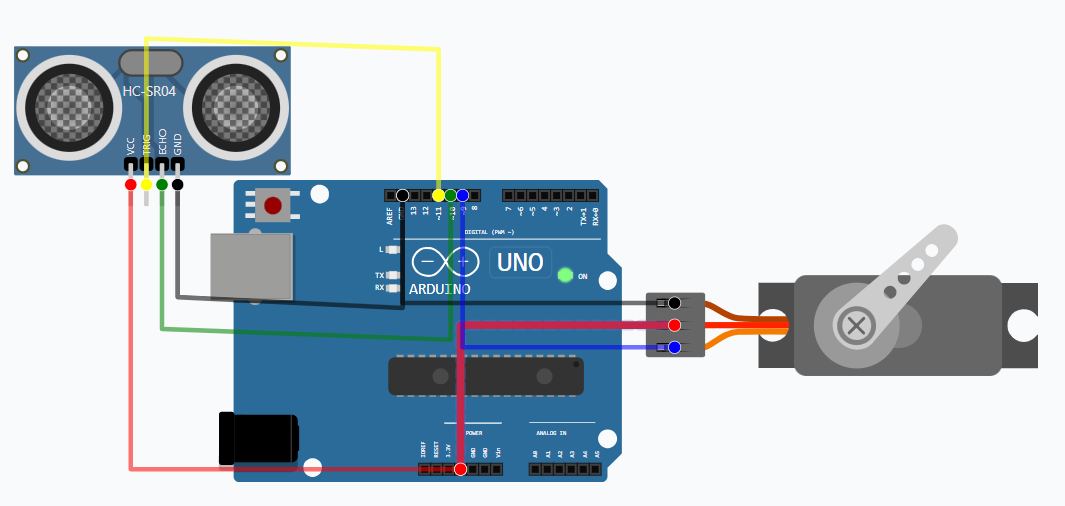}
    \caption{Circuit Schematic of a simple radar system generated by the visualization engine}
    \label{fig:schematic}
\end{figure}

\begin{figure}[H]
    \centering
    \includegraphics[width=\columnwidth]{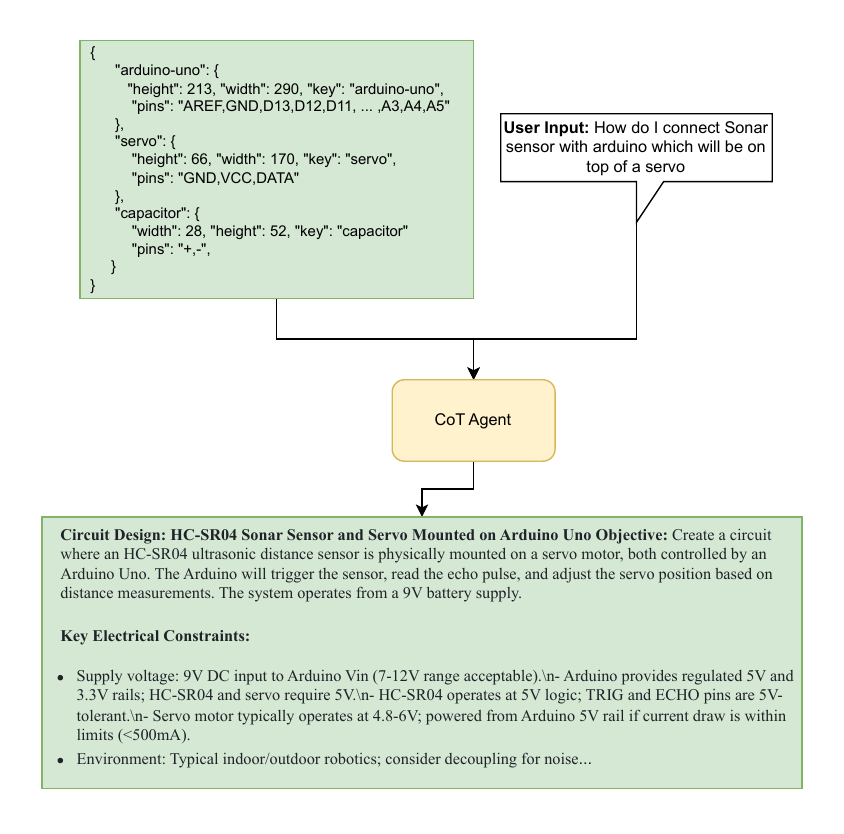}
    \caption{Workflow of the chain of thought agent, it takes the user's prompt and the retrieved components list and generates a reasoning for building the circuit }
    \label{fig:cot}
\end{figure}

\begin{table}[t]
\centering
\footnotesize % Natively scales the table down to fit the column
\setlength{\tabcolsep}{4pt} % Tightens the space between columns
\begin{threeparttable}
\caption{Performance of LLM-as-Judge Across Prompt Difficulty Levels (Percentage) After Removing the Chain-of-Thought Agent}
\label{tab:difficulty_breakdown_cot}
\begin{tabular}{@{} l c c c c @{}} % @{} removes invisible outer margins
\toprule
\textbf{Model} & \textbf{Easy (\%)} & \textbf{Medium (\%)} & \textbf{Hard (\%)} & \textbf{Overall (\%)} \\
\midrule
Gemini 2.5 Flash & 72.46 & 58.00 & 20.00 & 51.65 \\
Deepseek v3.1    & 55.00 & 37.00 & 29.00 & 38.83 \\
Qwen-3 235B      & 63.45 & 52.66 & 17.00 & 36.86 \\
GPT-5 mini       & 44.00 & 32.00 & 14.00 & 26.00 \\
Llama-3.3 70B    & 44.26 & 36.33 &  2.00 & 21.00 \\
\bottomrule
\end{tabular}
\begin{tablenotes}[flushleft] % flushleft forces proper paragraph wrapping
\item Note: Pass@1 scores are averaged across all runs. For each difficulty level in each run, the score is calculated as $100 \times (\text{Passed} \div \text{Total})$, where Total is the number of prompts at that difficulty.
\end{tablenotes}
\end{threeparttable}
\end{table}

\begin{table}[t]
\centering
\footnotesize % Keeps text readable while fitting 7 columns
\setlength{\tabcolsep}{3pt} % Tightens space between columns
\begin{threeparttable}
\caption{Performance Breakdown by Prompt Difficulty (Percentage) for both ERC and LLM-based evaluation}
\label{tab:difficulty_breakdown}
\begin{tabular}{@{} l c c c c c c @{}} % @{} removes invisible outer margins
\toprule
\multirow{2}{*}{\textbf{Model}} & 
\multicolumn{2}{c}{\textbf{Easy (\%)}} & 
\multicolumn{2}{c}{\textbf{Medium (\%)}} & 
\multicolumn{2}{c}{\textbf{Hard (\%)}} \\
% \cmidrule(lr) adds a trimmed line to group the sub-columns visually
\cmidrule(lr){2-3} \cmidrule(lr){4-5} \cmidrule(lr){6-7}
& \textbf{ERC} & \textbf{LLM} & \textbf{ERC} & \textbf{LLM} & \textbf{ERC} & \textbf{LLM} \\
\midrule
Gemini 2.5 Flash & 90.00 & 75.00 & 80.00 & 60.00 & 92.00 & 40.00 \\
Deepseek v3.1    & 95.00 & 55.00 & 80.00 & 36.67 & 86.00 & 36.00 \\
Qwen-3 235B      & 90.00 & 65.00 & 80.00 & 53.33 & 90.00 & 18.00 \\
GPT-5 mini       & 85.00 & 45.00 & 80.00 & 30.00 & 84.00 & 10.00 \\
Llama-3.3 70B    & 85.00 & 45.00 & 76.67 & 36.67 & 88.00 &  2.00 \\
\bottomrule
\end{tabular}
\begin{tablenotes}[flushleft]
\item Note: Pass@1 scores are averaged across all runs. For each difficulty level in each run, the score is calculated as $100 \times (\text{Passed} \div \text{Total})$, where Total is the number of prompts at that difficulty.
\end{tablenotes}
\end{threeparttable}
\end{table}

\begin{figure}[H]
    \centering
    \includegraphics[width=\columnwidth]{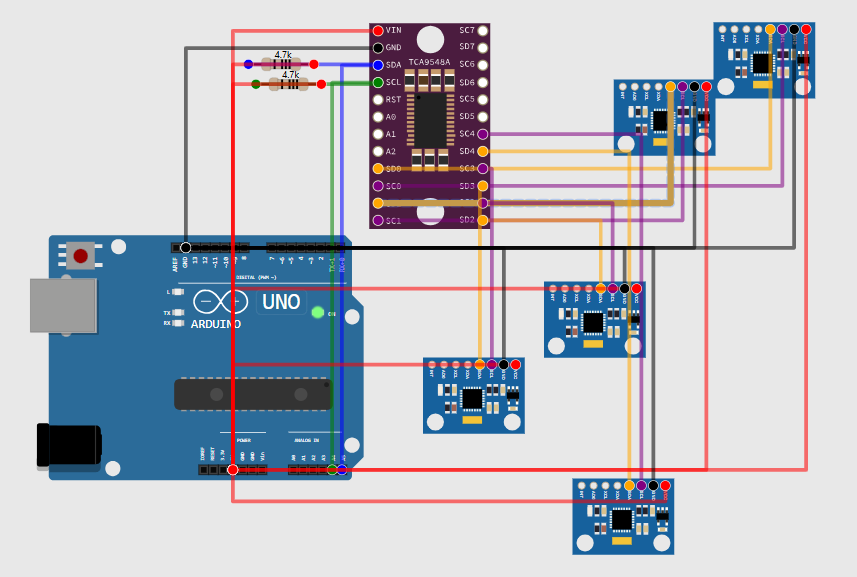}
    \caption{Circuit Schematic of Sign language glove using 5 IMU sensors for 5 fingertips generated by the visualization engine}
    \label{fig:schematic2}
\end{figure}

\begin{figure}[H]
    \centering
    \includegraphics[width=0.9\columnwidth]{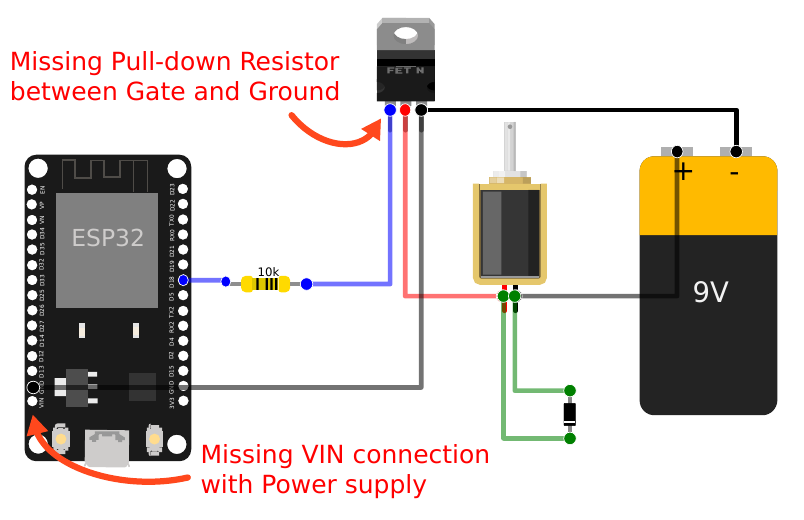}
    \caption{LLM-as-Judge evaluation of a CircuitLM-generated schematic for a solenoid driver. For the prompt ``How do I wire an N-channel MOSFET to drive a solenoid using PWM signal?", the framework generates a structurally compliant baseline using an ESP32 and an N-channel FET. However, the QA Agent correctly flags practical electrical faults, specifically identifying a missing pull-down resistor between the gate and ground and a missing VIN connection with the power supply.}
    \label{fig:eval_circuit}
\end{figure}

% \vfill\eject

\begin{lstlisting}[
  caption={\texttt{CircuitJSON} Schema Definition. The top, left, rotate are\\for placement of components, attrs is a flexible map for additional\\information, and parametric values.},
  label=lst:schema,
  numbers=left,          % Ensures numbers are on the left
  xleftmargin=2em,       % Pushes the listing right to make room for numbers
  frame=single,          % Optional: adds a box around it to see the boundaries clearly
  resetmargins=true,     % Resets the margins to align with the column
  breaklines=true        % Ensures long lines don't bleed into the next column
]
CircuitJSON
  version : Number
  author  : String
  parts : List of Part
  connections : List of Connection

Part
  type  : String   // e.g. "arduino-uno"
  id    : String   // unique instance ID
  top   : Number   // Y-coordinate
  left  : Number   // X-coordinate
  attrs : Map<String, Any>
  rotate: Number (optional)

Connection
  startPin: String  // e.g. "arduino:5V"
  endPin  : String  // e.g. "l298n:5V"
  color   : String  // wire color
  route   : List<String>  // routing instructions
\end{lstlisting}

\begin{table*}[t]
\centering
\small 
\caption{Qualitative Comparison of CircuitLM with Recent LLM-Based Hardware Generation Frameworks}
\label{tab:related_works_comparison}

% Define clean, readable colors for the check and cross marks
\newcommand{\cmark}{\textcolor{green!60!black}{\ding{51}}}
\newcommand{\xmark}{\textcolor{red}{\ding{55}}}

% Force tabularx to use vertically centered 'm' columns instead of top-aligned 'p' columns
\renewcommand{\tabularxcolumn}[1]{m{#1}}
% Custom column type: vertically AND horizontally centered
\newcolumntype{C}{>{\centering\arraybackslash}X}

\begin{tabularx}{\textwidth}{@{} l C C C C C C @{}}
\toprule
\textbf{Framework} & \textbf{Target Domain} & \textbf{Input Modality} & \textbf{Output Format} & \textbf{Component Retrieval} & \textbf{Automated Rule Checking} & \textbf{Visual Rendering} \\
\midrule

VerilogEval \cite{thakur2023verilogeval} & Digital RTL / Logic & Natural Language & Verilog & \xmark & Syntax / Compilation & \xmark \\
\addlinespace

LLM4Netlist \cite{ye2025llm4netlist} & Digital Logic Gates & Natural Language & SPICE Netlist & \xmark & Simulation-based & \xmark \\
\addlinespace

Schemato \cite{matsuo2025schematollmnetlisttoschematic} & Analog Circuits & Netlist & Schematic & N/A & \xmark & \cmark \\
\addlinespace

EESchematic \cite{liu2025eeschematicmultimodalllmbasedai} & Analog Circuits & Multimodal (Text/Image) & Schematic & \xmark & \xmark & \cmark \\
\addlinespace

MenTeR \cite{chen2025menterfullyautomatedmultiagentworkflow} & RF / Analog & Natural Language & Netlist / Layout & \hspace{5mm}\cmark \newline(Diagram-aware) & Simulation-based & \xmark \\
\midrule

\rowcolor{gray!10}
\textbf{CircuitLM (Ours)} & \textbf{Embedded Systems} & \textbf{Natural Language} & \textbf{CircuitJSON} & \hspace{5mm}\textbf{\cmark} \newline \textbf{(Components dataset)} & \hspace{5mm}\textbf{\cmark} \newline \textbf{(ERC \& LLM)} & \hspace{5mm}\textbf{\cmark} \newline \textbf{(Interactive)} \\

\bottomrule
\end{tabularx}
\vspace{-2mm}
\end{table*}

\end{document}